\documentclass[runningheads,a4paper]{llncs}
\usepackage[top=2.5cm, bottom=2.5cm, left=2.5cm, right=2.5cm]{geometry}
\usepackage{amsmath}
\usepackage{amssymb}
\usepackage{graphicx}
\usepackage{booktabs}
\usepackage{listings}
\usepackage{xcolor}
\usepackage{hyperref}
\usepackage{caption}
\usepackage{subcaption}

\usepackage{url}
\usepackage{color}
\usepackage{cite}
\usepackage{epsfig}
\usepackage{microtype}
\usepackage{hyperref}

\usepackage[nomargin,inline,marginclue,draft]{fixme}
\usepackage{cleveref}
\fxsetup{status=draft}
\fxsetup{theme=color, mode=multiuser} \FXRegisterAuthor{me}{ame}{\color{red}Me}
\newcommand{\orcid}[1]{\href{https://orcid.org/#1}{\includegraphics[scale=0.02]{figures/ORCIDiD_icon128x128.jpg}}} 

\hypersetup{
    colorlinks=true
}
\title{The 2024 Brain Tumor Segmentation (BraTS) Challenge: Glioma Segmentation on Post-treatment MRI}
\titlerunning{2024 BraTS Post-Treatment Glioma Challenge}

\author{
Maria Correia de Verdier\inst{1,2,*,\dag,\ddag,\S,\P}
Rachit Saluja\inst{3,4,*,\dag,\ddag}
Louis   Gagnon\inst{2,7,\dag,\S,\P}
Dominic  LaBella\inst{8,\dag,\ddag,\S,\P}
Ujjwall  Baid\inst{5,\dag,\ddag,\P}
Nourel Hoda Tahon\inst{9,\dag,\S,\P}
Martha  Foltyn-Dumitru\inst{6,\ddag,\P}
Jikai  Zhang\inst{10,\S,\P}
Maram  Alafif\inst{2,\S}
Saif  Baig\inst{11,\S}
Ken  Chang\inst{12,\S}
Gennaro  D'Anna\inst{13,\S}
Lisa  Deptula\inst{14,\S}
Diviya  Gupta\inst{15,\S}
Muhammad Ammar Haider\inst{16,\S}
Ali  Hussain\inst{17,\S}
Michael  Iv\inst{12,\S}
Marinos  Kontzialis\inst{18,\S}
Paul   Manning\inst{2,19,\S}
Farzan  Moodi\inst{20,\S}
Teresa  Nunes\inst{21,22,\S}
Aaron   Simon\inst{23,\S}
Nico  Sollmann\inst{24,25,\S}
David  Vu\inst{26,\S}
Maruf Adewole\inst{27,\ddag}
Jake Albrecht\inst{28,\ddag}
Udunna  Anazodo\inst{29,\ddag}
Rongrong Chai\inst{28,\ddag}
Verena Chung\inst{28,\ddag}
Shahriar  Faghani\inst{30,\ddag}
Keyvan Farahani\inst{31,\ddag}
Anahita Fathi Kazerooni\inst{32,33,\ddag}
Eugenio Iglesias\inst{34,\ddag}
Florian Kofler\inst{35,\ddag}
Hongwei Li\inst{34,\ddag}
Marius George  Linguraru\inst{36,37,\ddag}
Bjoern  Menze\inst{38,\ddag}
Ahmed W. Moawad\inst{39,\ddag}
Yury Velichko\inst{40,\ddag}
Benedikt  Wiestler \inst{41,\ddag}
Talissa  Altes\inst{9,\P}
Patil Basavasagar\inst{42,\P}
Martin Bendszus\inst{6,\P}
Gianluca Brugnara\inst{6,\P}
Jaeyoung Cho\inst{6,\P}
Yaseen  Dhemesh\inst{9,\P}
Brandon K.K. Fields\inst{43,\S,\P}
Filip  Garrett\inst{44,\P}
Jaime  Gass\inst{9,\P}
Lubomir Hadjiiski\inst{42,\P}
Jona  Hattangadi-Gluth\inst{45,\P}
Christopher Hess\inst{43,\P}
Jessica L.  Houk\inst{46,\P}
Edvin  Isufi\inst{9,\P}
Lester J.  Layfield\inst{44,\P}
George Mastorakos\inst{47,\P}
John Mongan\inst{43,\P}
Pierre Nedelec\inst{43,\P}
Uyen Nguyen\inst{48,\P}
Sebastian Oliva\inst{42,\P}
Matthew W.  Pease\inst{49,\ddag,\P}
Aditya Rastogi\inst{6,\P}
Jason  Sinclair\inst{44,\P}
Robert X. Smith\inst{47,\P}
Leo P. Sugrue\inst{43,\P}
Jonathan  Thacker\inst{9,\P}
Igor Vidic\inst{47,\P}
Javier  Villanueva-Meyer\inst{43,\P}
Nathan S. White\inst{47,\P}
Mariam  Aboian\inst{50,\dag,\ddag}
Gian Marco Conte\inst{30,\dag,\ddag}
Anders Dale\inst{2,\P}
Mert R.  Sabuncu\inst{3,4,\dag}
Tyler M. Seibert\inst{51,\S,\P}
Brent  Weinberg\inst{52,\ddag,\S}
Aly  Abayazeed\inst{53,\dag,\ddag,\S}
Raymond  Huang\inst{54,55,\dag,\ddag,\S}
Sevcan Turk\inst{42,\P}
Andreas M. Rauschecker\inst{43,\S,\P}
Nikdokht  Farid\inst{2,\dag,\P}
Philipp  Vollmuth\inst{6,56,\dag,\ddag,\P}
Ayman  Nada\inst{9,\dag,\S,\P}
Spyridon  Bakas\inst{5,\dag,\ddag,\P}
Evan  Calabrese\inst{46,\dag,\ddag,\S,\P}
Jeffrey D.  Rudie\inst{2,26, +,\dag,\ddag,\S,\P}
}

\authorrunning{Correia de Verdier, Saluja \& Rudie}

\institute{\scriptsize{
Department of Surgical Sciences, section of Neuroradiology, Uppsala University, Sweden \and
Department of Radiology, University of California San Diego, CA, USA \and
Department of Electical and Computer Engineering , Cornell University and Cornell Tech, New York, NY, USA \and
Department of Radiology, Weill Cornell Medicine \and
Department of Pathology and Laboratory Medicine, Indiana Univeristy, IN, USA \and
Department of Neuroradiology, Heidelberg University, Germany \and
University of Laval, Quebec City, QC, Canada \and
Department of Radiation Oncology, Duke University, Durham, NC, USA \and
Radiology Department, University of Missouri, Columbia, MO, USA \and
Department of Electrical and Computer Engineering, Duke University, Durham, NC, USA \and
Department of Radiology, Nassau Medical Center, East Meadow, NY, USA \and
Department of Radiology, Stanford University, CA, USA \and
Neuroimaging Unit, ASST Ovest Milanese - Legnano (Milan) - Italy \and
Ross University School of Medicine, Bridgetown, Barbados \and
Department of Radiology, University of California Los Angeles, CA, USA \and
C.M.H. Lahore Medical College \& Institute of Dentistry, Lahore, Pakistan \and
Department of Imaging Sciences, University of Rochester Medical Center, Rochester, NY, USA \and
Department of Radiology, Department of Radiology, Northwestern University, Chicago, IL, USA \and
Department of Radiology,  VA San Diego Medical Center \and
School of Medicine, Iran University of Medical Sciences, Tehran, Iran \and
Neuroradiology Department, Hospital Garcia de Orta - Unidade Local de Saúde de Almada Seixal, EPE; Almada, Portugal \and
Imaging Department, Hospital da Luz Lisboa, Lisbon, Portugal \and
Department of Radiation Oncology, University of California Irvine, CA, USA \and
Department of Diagnostic and Interventional Radiology, University Hospital Ulm, Ulm, Germany \and
Department of Diagnostic and Interventional Neuroradiology, School of Medicine, Klinikum rechts der Isar, Technical University of Munich, Munich, Germany \and
Department of Radiology, Scripps Clinic Medical Group, CA, USA \and
Medical Artificial Intelligence Laboratory (MAI Lab), Lagos, Nigeria \and
Sage Bionetworks \and
Montreal Neurological Institute (MNI), McGill University, Montreal, QC, Canada \and
Department of Radiology, Mayo Clinic, Rochester, MN, USA \and
National Institute of Health, USA \and
Division of Neurosurgery, The Children's Hospital of Philadelphia, PA, USA \and
Department of Neurosurgery, University of Pennsylvania, Philadelphia, PA, USA \and
Athinoula A Martinos Center for Biomedical Imaging, Massachusetts General Hospital, Boston, MA, USA \and
Helmholtz AI, Helmholtz Munich, Germany \and
Children’s National Hospital, Washington DC, USA  \and
George Washington University, Washington DC, USA \and
University of Zurich, Switzerland \and
Department of Radiology, Mercy Catholic Medical Center, Darby, PA, USA \and
Department of Radiology, Northwestern University, Chicago IL, USA \and
Klinikum rechts der Isar, Technical University of Munich, Germany \and
Department of Radiology , University of Michigan, MI, USA \and
Department of Radiology and Biomedical Imaging, University of California San Francisco, CA, USA \and
Pathology Department, University of Missouri, Columbia, MO, USA \and
Department of Radiation Oncology, University of California San Diego, CA, USA \and
Department of Radiology, Duke University, Durham, NC, USA \and
Cortechs.ai \and
School of Medicine, University of California San Diego, CA, USA \and
Department of Neurosurgery, Indiana Univeristy, IN, USA \and
Department of Radiology, The Children's Hospital of Philadelphia, PA, USA \and
Departments of Radiation Medicine and Applied Sciences, Radiology, and Bioengineering, University of California San Diego, CA, USA \and
Department of Radiology and Imaging Sciences, Emory University, Atlanta, GA, USA \and
Stanford University, CA, USA \and
Department of Radiology, Brigham and Women's Hospital, Boston, MA, USA \and
Harvard Medical School, Boston, MA, USA \and
Division for Computational Radiology \& Clinical AI, University Hospital Bonn
}
\\
\vspace{10pt} 
\textsuperscript{*} Equal first authors.\\
\textsuperscript{+} Senior author.\\
\textsuperscript{\dag} 2024 BraTS Post-Treatment Challenge Organizer. \\
\textsuperscript{\ddag} 2024 BraTS Organizer.\\
\textsuperscript{\S} Annotator.\\ 
\textsuperscript{\P} Data Contributor.\\ 
\textsuperscript{**} Corresponding author: \email{\{jeff.rudie@gmail.com\}}}

\begin{document}
    \mainmatter
    \maketitle
    \setcounter{footnote}{0} 
    \begin{abstract}
        Gliomas are the most common malignant primary brain tumors in adults and one of the deadliest types of cancer. There are many challenges in treatment and monitoring due to the genetic diversity and high intrinsic heterogeneity in appearance, shape, histology, and treatment response. Treatments include surgery, radiation, and systemic therapies, with magnetic resonance imaging (MRI) playing a key role in treatment planning and post-treatment longitudinal assessment. The 2024 Brain Tumor Segmentation (BraTS) challenge on post-treatment glioma MRI will provide a community standard and benchmark for state-of-the-art automated segmentation models based on the largest expert-annotated post-treatment glioma MRI dataset. Challenge competitors will develop automated segmentation models to predict four distinct tumor sub-regions consisting of enhancing tissue (ET), surrounding non-enhancing T2/fluid-attenuated inversion recovery (FLAIR) hyperintensity (SNFH), non-enhancing tumor core (NETC), and resection cavity (RC). Models will be evaluated on separate validation and test datasets using standardized performance metrics utilized across the BraTS 2024 cluster of challenges, including lesion-wise Dice Similarity Coefficient and Hausdorff Distance. Models developed during this challenge will advance the field of automated MRI segmentation and contribute to their integration into clinical practice, ultimately enhancing patient care.
    \end{abstract}
    
    
    \section{Introduction}

Gliomas are among the deadliest types of cancer and constitute the most prevalent malignant primary brain tumors in adults. According to the Central Brain Tumor Registry of the United States, gliomas as a group represent approximately 25\% of all primary brain tumors and 80\% of malignant primary brain and central nervous system tumors, with diffuse gliomas being the most common malignant subtype \cite{low2022primary}. Diffuse gliomas are a genetically diverse group of disorders, comprising astrocytoma, oligodendroglioma, and glioblastoma, encompassing both low- and high-grade gliomas (World Health Organization [WHO] grades 2-4) \cite{louis2016classification}. They are characterized by their infiltrative growth pattern within the central nervous system, presenting substantial challenges for treatment and monitoring due to their variability in biological behavior, prognosis, and response to therapy.

\

Treatment for diffuse gliomas involves a multi-modal approach tailored to the tumor’s characteristics and the patient’s health and includes surgery, radiation therapy, and systemic therapies (i.e. chemotherapy, immunotherapy, and targeted therapy). Years of extensive research to improve diagnosis, characterization, and treatment have decreased mortality rates, but gliomas remain one of the deadliest types of malignancies. The prognosis varies depending on tumor grade and molecular type, the patient’s age and health, and how they respond to treatment. Median survival is lowest for glioblastoma (isocitrate dehydrogenase [IDH]-wildtype, WHO grade 4), with overall survival of 8 months, extending to 14 months with standard-of-care treatment, while it is highest for oligodendroglioma at approximately 17 years \cite{low2022primary, stupp2005radiotherapy}. Magnetic resonance imaging (MRI) remains the gold standard for post-treatment imaging across the spectrum of diffuse gliomas. It provides crucial information on tumor size, location, and morphological changes over time. Post-treatment imaging of diffuse gliomas is a fundamental part of patient management that significantly influences clinical decision making and outcomes. 

\

Most prior research on brain tumor segmentation has been performed in the pre-treatment setting, with a recent survey finding that 98.3\% of published glioma segmentation studies had been performed on pre-operative imaging \cite{hoebel2023expert}. These studies typically utilized pre-operative data from The Cancer Imaging Archives (TCIA) \cite{clark2013cancer} or prior Brain Tumor Segmentation (BraTS) challenges \cite{menze2014multimodal, bakas2018identifying, baid2021rsna}, which first began in 2012.

\

The combination of treatment-related changes, including resection cavities, blood products, post-radiation inflammation, and gliosis, combined with the already naturally ill-defined tumor borders seen in infiltrative diffuse gliomas makes segmentation a challenging task, which is not addressed by segmentation models trained on untreated tumors. While several recent studies involving glioma segmentation have been performed in the post-treatment setting \cite{kickingereder2019automated, chang2019automatic, rudie2022longitudinal, lotan2022development, ghaffari2022automated, sorensen2023evaluation, vollmuth2023artificial, 10.1093/neuonc/noac209.1145, fieldsUCSF}, relatively little post-treatment glioma data with voxelwise annotations exists publicly \cite{fieldsUCSF, cepeda2023rio}. 

\

The 2024 BraTS challenge is the first to focus on post-treatment gliomas. The goal is to create a public database of annotated post-treatment glioma MRIs and a benchmarking environment for the development and evaluation of deep learning segmentation algorithms. The 2024 BraTS post-treatment glioma challenge utilizes a new dataset consisting only of post-treatment MRIs of low- and high-grade diffuse gliomas. It includes the resection cavity (RC) as a new tissue subregion. Algorithms developed from this challenge could be used as an objective measure for assessing residual tumor volume, which can guide treatment. Moreover, the dataset can serve as a starting point for future studies aimed at distinguishing treatment changes from residual/recurrent tumor, predicting outcomes, and evaluating response to treatment.

    \section{Methods}
        \subsection{Data}

\begin{figure}[t]
      \centering
      \includegraphics[width=1\linewidth]{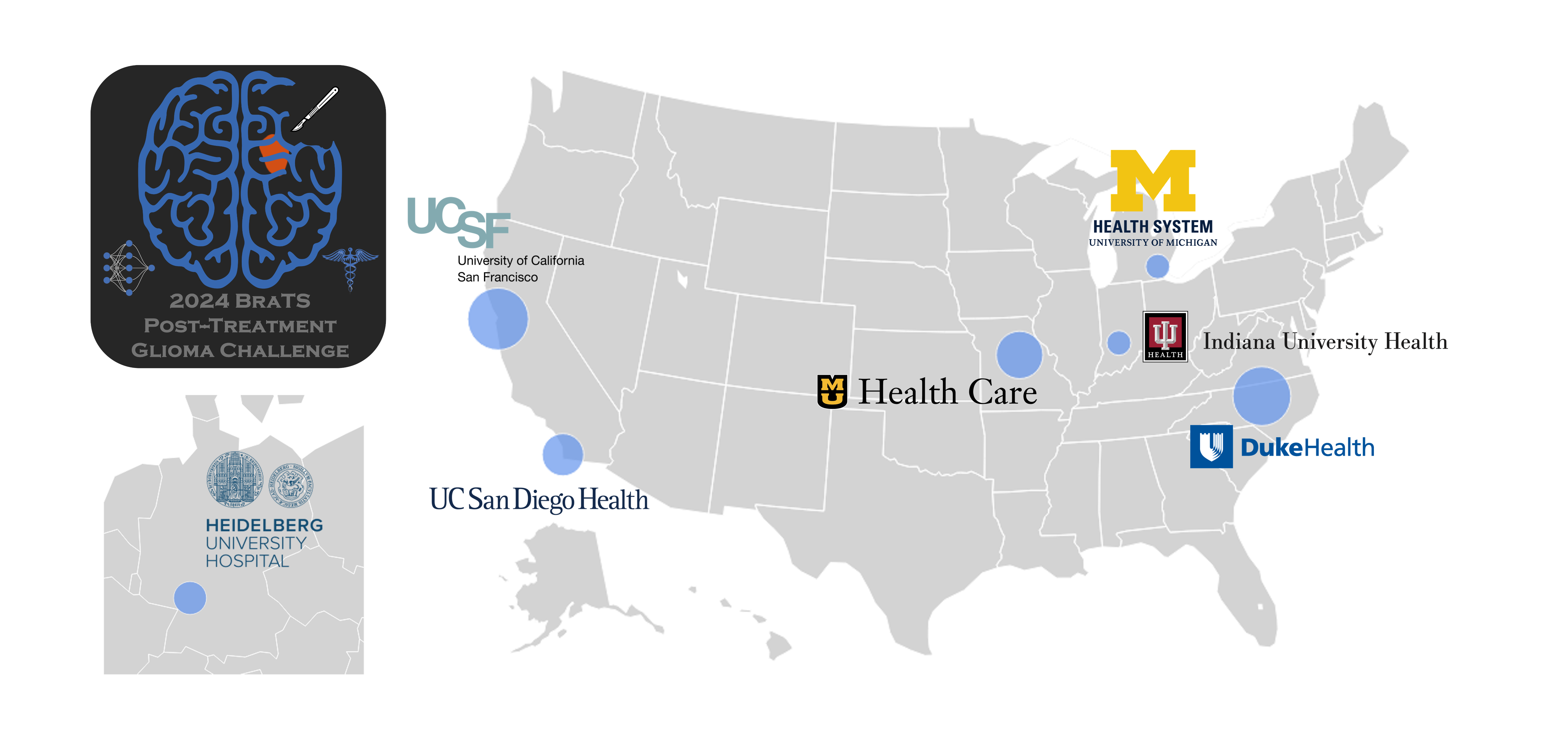}  
      \caption{Map showing institutions from USA and Germany contributing data to the 2024 Brain Tumor Segmentation (BraTS) post-treatment glioma challenge and their relative sample size.}
    \label{figure: Fig1-BubbleMap}
\end{figure}

Retrospective multi-institutional cohorts of patients, diagnosed with diffuse gliomas having already undergone treatment, which may include surgery, radiation, and/or systemic therapy. Data was contributed from seven different academic medical centers (Figure \ref{figure: Fig1-BubbleMap}, Table \ref{tab:cases}): Duke University, University of California San Francisco (UCSF; which consists of the previously described UCSF Longitudinal Post-Treatment Diffuse Glioma [UCSF-LPTDG]) \cite{fieldsUCSF}, University of Missouri Columbia, University of California San Diego, Heidelberg University Hospital, University of Michigan and Indiana University. 

\begin{table}[h!]
\centering
\captionsetup{skip=10pt}
\begin{tabular}{lc} \toprule
\textbf{Contributing Sites} & \textbf{Number of Cases (approximate)} \\ \midrule
Duke University                  & 680                                    \\
University of California San Francisco & 600                      \\
University of Missouri Columbia        & 400                      \\
University of California San Diego      & 350                      \\
Heidelberg University Hospital          & 300                      \\
University of Michigan                  & 100                      \\
Indiana University                      & 70                       \\ \midrule
\textbf{Total}                          & \textbf{2200}             \\ \bottomrule
\end{tabular}
\caption{Number of cases contributed by each site.}
\label{tab:cases}
\end{table}

The patients have been clinically scanned with multiparametric MRI (mpMRI) acquisition protocols including the following MRI sequences:

\begin{enumerate}
    \item Pre-contrast T1-weighted (T1)
    \item Contrast-enhanced T1-weighted (T1-Gd)
    \item T2-weighted (T2)
    \item T2-weighted fluid-attenuated inversion recovery (FLAIR)

\end{enumerate}

\subsection{Data Preprocessing}

\begin{figure}[t]
      \centering
      \includegraphics[width=1\linewidth]{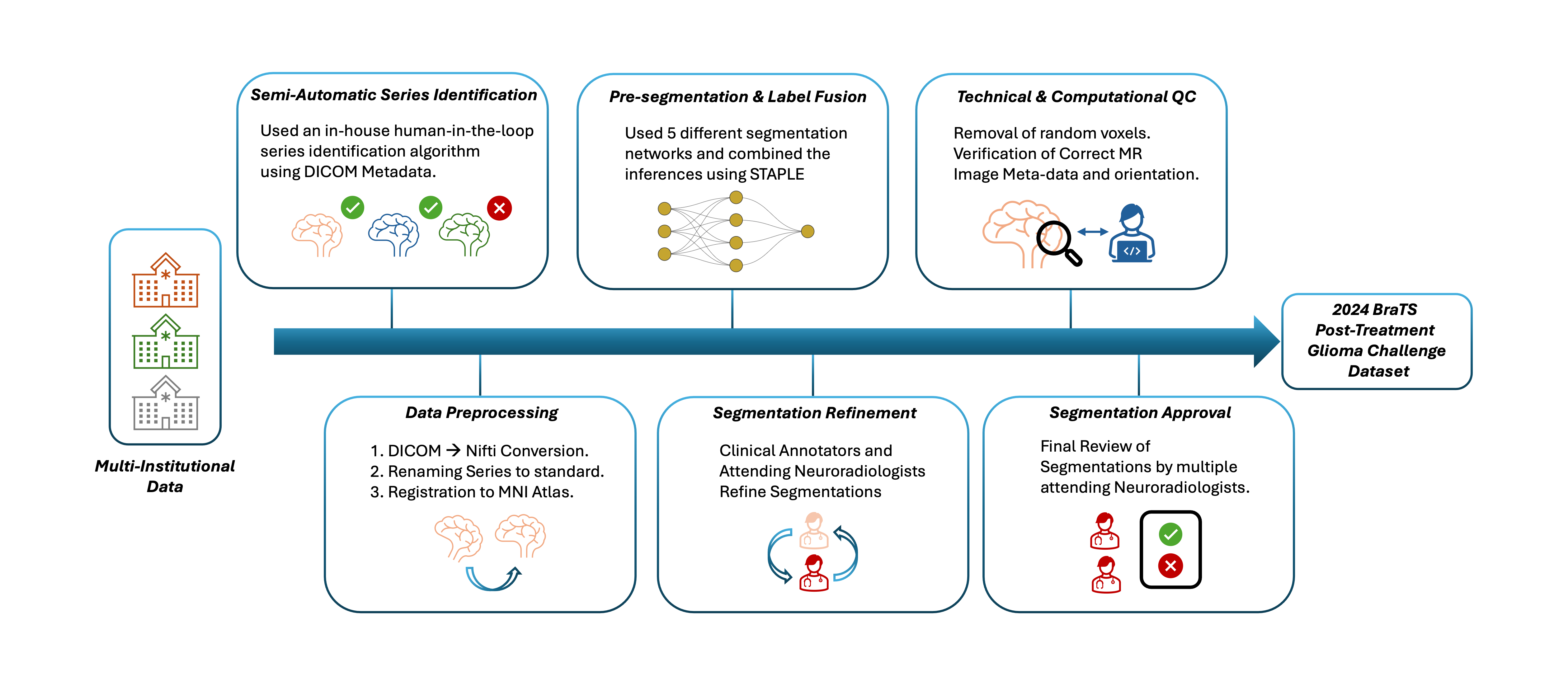}  
      \caption{Data Processing and Annotation Workflow for creating the 2024 BraTS post-treatment glioma challenge dataset.}
    \label{figure: Fig2-Workflow}
\end{figure}

The preprocessing pipeline (Figure \ref{figure: Fig2-Workflow}) applied to all the data considered in the 2024 BraTS post-treatment glioma challenge was similar to the one evaluated and followed by the BraTS 2017-2023 challenges \cite{menze2014multimodal, bakas2018identifying, baid2021rsna, bakas2017advancing}. Raw MRI scans were first carefully reviewed by radiologists from individual institutions. The T1, T1-Gd, T2, and FLAIR sequences were then extracted and named according to the standard BraTS naming convention. The dcm2niix software was applied to the four sequences to convert the raw scans from their original Digital Imaging and Communications in Medicine (DICOM) file format to the Neuroimaging Informatics Technology Initiative (NIfTI) file format \cite{cox2004sort}. Following the conversion to NIfTI files, we performed brain extraction to remove any apparent non-brain tissue (e.g., neck fat, skull, eyeballs) using HD-BET\footnote{\href{https://github.com/CCI-Bonn/HD-BET}{https://github.com/CCI-Bonn/HD-BET}} \cite{isensee2019automated}. The brain-extracted T1, T1-Gd, T2, and FLAIR sequences were registered to the Linear Symmetrical MNI Atlas using affine registration via the CapTK/Greedy software \cite{pati2020cancer}. Several institutions that contributed to the dataset processed their data in a similar in-house pipeline or the FeTS 2.0 platform\footnote{\href{https://fets-ai.github.io/FL-PoST/}{https://fets-ai.github.io/FL-PoST/}}.

\

The four sequences, or the image volume data, were then segmented using five different pre-segmentation approaches, all of which used the nnU-Net \cite{isensee2021nnu}.
            
\begin{enumerate}
    \item 3D U-Net trained by nnU-Net V1 on post-treatment glioma data acquired by Duke and UCSF using five-fold cross-validation, which is the Federated Learning for Postoperative Segmentation of Treated glioblastoma (FL-PoST) pre-segmentation method.
    \item 3D U-Net trained by nnU-Net V2 on post-treatment glioma data acquired by Duke and UCSF using five-fold cross-validation. 
    \item 3D SegResNet\cite{myronenko2022automated} trained using MONAI\footnote{\href{https://docs.monai.io/en/stable/networks.html\#segresnet-block}{MONAI SegResNet Block}}(Medical Open Network for AI) and nn-UNet V2 on post-treatment glioma data acquired by Duke and UCSF using five-fold cross-validation.
    \item A 3D proprietary nnU-Net trained by Cortechs.ai (Neuroquant Glioma\textsuperscript{TM}) on pre- and post-treatment glioma data from various datasets from the cancer imaging archives (TCIA)\cite{10.1093/neuonc/noac209.1145}.
    \item A 3D nnU-Net trained on the UCSF-LPTDG dataset and various TCIA post-treatment glioma datasets annotated by Cortechs.ai.

\end{enumerate}    

The five segmentations produced by these networks were combined using the Simultaneous Truth and Performance Level Estimation (STAPLE) fusion algorithm \cite{rohlfing2004performance}. In addition, digital subtraction images were created between the T1-Gd and T1 sequences to facilitate the annotation process for radiologists in cases of more subtle enhancement or confounding areas of T1 intrinsic hyperintensity. This data was handed off to radiologists to make the annotation process easier and more efficient.

\subsection{Tumor Annotation Protocol}

\begin{figure}[t]
      \centering
      \includegraphics[width=1\linewidth]{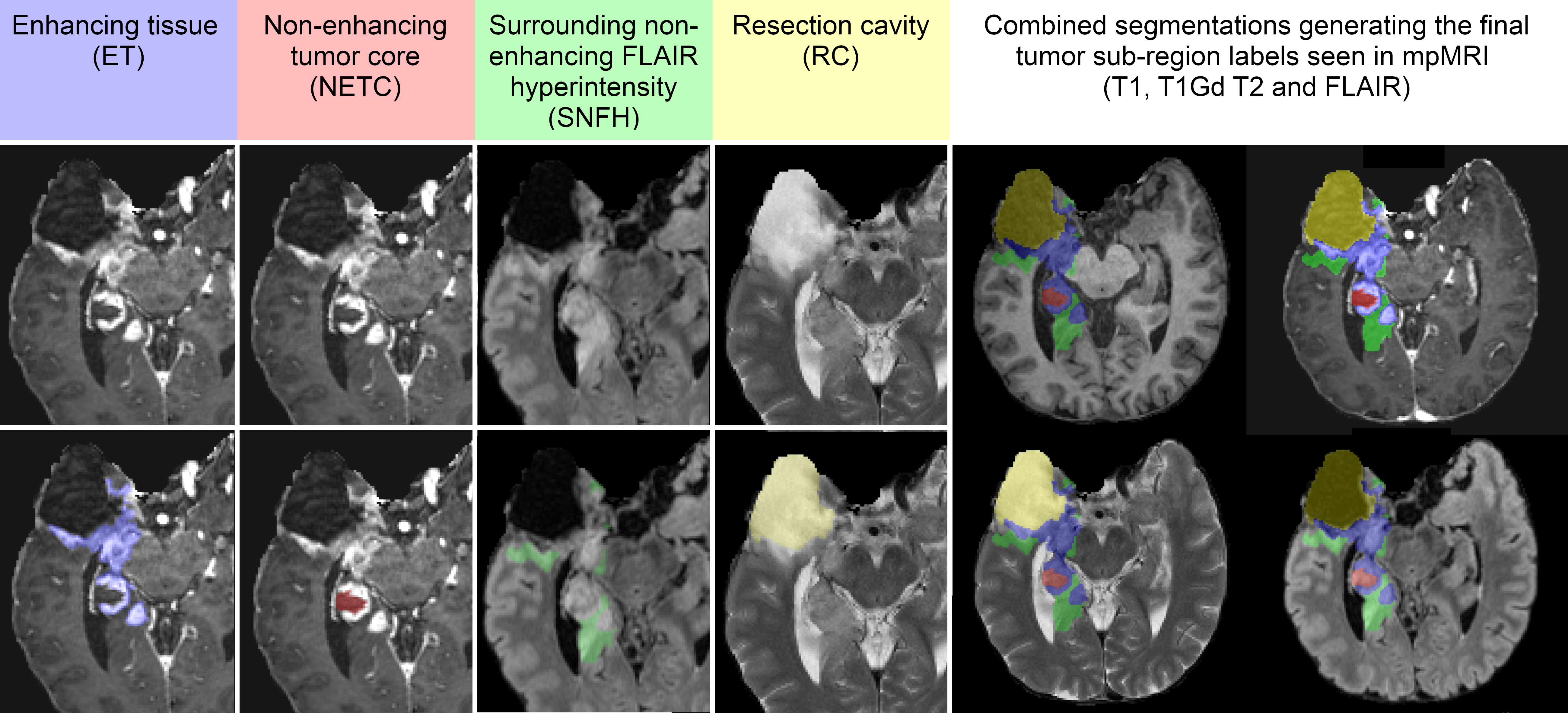}  
      \caption{Tumor sub-regions considered in the 2024 BraTS post-treatment glioma challenge. Image panels with the tumor sub-regions annotated in the different mpMRI scans and combined segmentations on mpMRI. The enhancing tissue (blue) visible on a T1-Gd scan, the non-enhancing tumor core (red) visible on a T1-Gd scan, the surrounding non-enhancing FLAIR hyperintensity (green) visible on a FLAIR scan and the resection cavity (yellow) visible on a T2 scan. The combined segmentations generating the final tumor sub-region labels visible on mpMRI, as provided to the challenge participants: enhancing tissue (blue), the surrounding non-enhancing FLAIR hyperintensity (green), the non-enhancing tumor core (red), and the resection cavity (yellow).}
    \label{figure: Fig3-Subregions}
\end{figure}

The data considered in the 2024 BraTS post-treatment glioma challenge was similar to the paradigm of the BraTS 2021-2023 challenge data \cite{menze2014multimodal, bakas2018identifying, baid2021rsna, bakas2017advancing}, though with modifications specific to the post-treatment setting. The annotation (Figure \ref{figure: Fig3-Subregions}) of these data followed a pre-defined clinically approved annotation protocol (defined by expert neuroradiologists and radiation oncologists). This annotation protocol was provided to all clinical annotators, describing in detail instructions on what the segmentations of each tumor sub-region should include with numerous examples of more challenging cases\footnote{\href{https://drive.google.com/file/d/10oI-KUxQVT0FpClOYZVu_zad1yi4WQC6/view}{Tumor Annotation Protocol Document}}. 

Summary of specific instructions:

\begin{enumerate}
    \item \textbf{Enhancing tissue (ET)}: This delineates the hyperintense signal on T1-Gd, after excluding the vessels. Any areas of thick or nodular enhancement are included in the ET class, though typical treatment-related thin linear enhancement along and within resection cavities and along the dura was not included in the ET class. A T1-Gd - T1 subtraction image was provided to help identify subtle areas of enhancement and to distinguish areas of intrinsic T1 hyperintensity from enhancement.
    \item \textbf{Non-enhancing tumor core (NETC)}: This outlines regions appearing dark on both T1 and T1-Gd images (denoting necrosis/cysts), and dark regions on T1-Gd that appear brighter on T1 and are not otherwise clearly represented by a prior RC. 
    \item \textbf{Surrounding non-enhancing FLAIR hyperintensity (SNFH)}: This tissue typically includes edema and infiltrating tumor. Given the post-treatment nature of the scans, any T2/FLAIR signal abnormalities, including radiation-related hyperintensity, gliosis, edema, and non-enhancing tumor, were included in the SNFH label. Symmetric or patchy white matter hyperintensities clearly related to chronic microvascular ischemic disease or periventricular capping are not included.
    \item \textbf{Resection cavity (RC)}: The RC class consists of both recent and chronic resection cavities. Chronic resection cavities, which are typically older than 3-6 months, were considered those with signal intensity isointense to cerebrospinal fluid on both T1 and T2/FLAIR images. More recent resection cavities often contained air, blood, and/or proteinaceous materials, exhibiting variable signal characteristics.
\end{enumerate}

The annotators were provided with the four mpMRI sequences as well as a T1 contrast subtraction (T1-Gd - T1) image and given the flexibility to use their tool of preference for making the segmentations, following a hybrid approach where the presegmentations were refined manually. Once the tumor segmentations were refined by the annotators, the approver reviewed the segmentations. Annotators spanned across various experience levels and clinical/academic ranks, while the approvers were board-certified neuroradiologists on the organizing committee with prior annotation experience. The approver would review the tumor segmentations in tandem with the corresponding MRI scans, and, if the segmentations were not of satisfactory quality, would send them back to the annotators for further refinement. This iterative approach was followed for all cases until their respective segmentations reached satisfactory quality for public availability and were noted as the final ground truth segmentation labels for these scans. The test data were annotated and approved by two sets of annotators/approvers to assess inter-rater reliability.

\begin{figure}[t]
      \centering
      \includegraphics[width=1\linewidth]{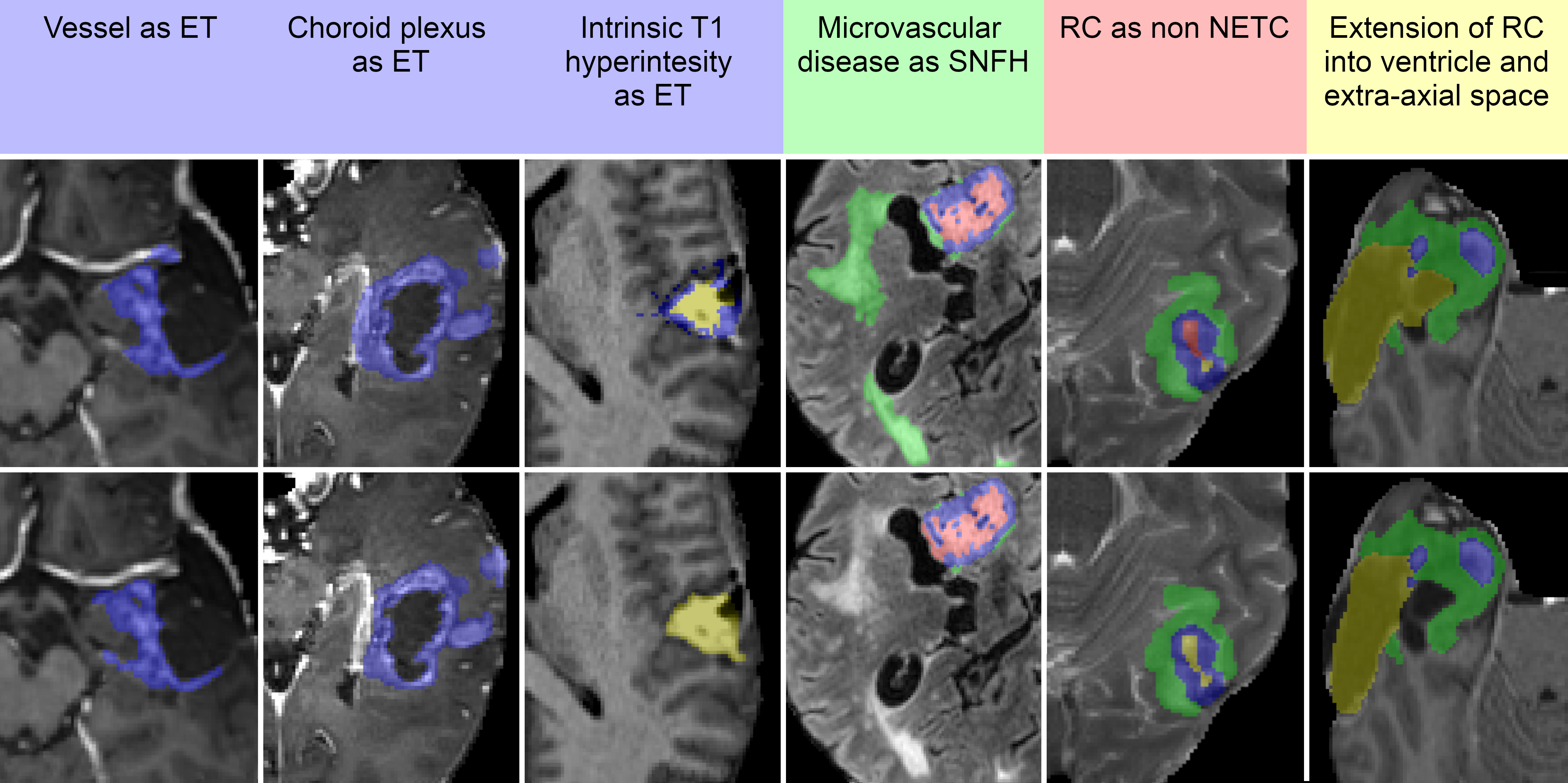}  
      \caption{Common errors in automated segmentations in the 2024 BraTS post-treatment glioma challenge. The top row shows typical segmentation errors, and the bottom row shows manually corrected labels. Color codes: blue for enhancing tissue (ET), red for non-enhancing tumor core (NETC), green for surrounding non-enhancing FLAIR hyperintensity (SNFH), and yellow for resection cavity (RC). Additional Detailed examples are available in the \href{https://drive.google.com/file/d/10oI-KUxQVT0FpClOYZVu_zad1yi4WQC6/view}{Tumor Annotation Protocol Document}.}
    \label{figure: Fig4-Errors}
\end{figure}

Common errors of automated segmentations (Figure \ref{figure: Fig4-Errors}): Based on observations, we have identified some common errors in automated segmentations. The most typical errors in the current challenge were:

\begin{enumerate}
    \item The segmentation of vessels and choroid plexus as ET. 
    \item The segmentation of non-enhancing lesions that have intrinsic T1 hyperintensity as ET. 
    \item Undersegmentation of SNFH or ET when subtle.
    \item The segmentation of white matter changes from microvascular disease unrelated to treatment as SNFH.
    \item The segmentation of RC as NETC or vice versa. 
    \item Extension of RC into ventricles or extra-axial spaces.
\end{enumerate}

        \section{Performance Evaluation}

The challenge is hosted on the \href{https://www.synapse.org/#!Synapse:syn53708249/wiki/627500}{Synapse Platform} (Sage Bionetworks). Following the paradigm of algorithmic evaluation in machine learning, the data included in the 2024 BraTS post-treatment glioma challenge were divided into training (70\%), validation (10\%), and testing datasets (20\%). Challenge participants receive ground truth labels exclusively for the training dataset, while the validation dataset is provided without any associated ground truth, and the testing dataset remains completely hidden. The evaluation metrics used in this challenge are the same metrics from previous BraTS 2023 challenges \cite{kazerooni2023brain, moawad2023brain, labella2023asnr, adewole2023brain, labella2024analysis, labella2024multi}.

\

In terms of evaluation metrics, we use:
\begin{enumerate}
    \item Lesion-wise Dice Similarity Coefficient (DSC), which measures voxelwise segmentation overlap between predicted and ground truth segmentations, ignoring true negative voxels.
    \item Lesion-wise 95\% Hausdorff distance (HD95), which measures the distance between the center of the predicted and ground truth segmentations.
\end{enumerate}

The lesion-wise DSC\footnote{\href{https://github.com/rachitsaluja/BraTS-2023-Metrics}{https://github.com/rachitsaluja/BraTS-2023-Metrics}} and lesion-wise HD95 metrics were designed to assess model performance at the level of individual lesions rather than across the entire image. This methodology prevents our evaluation from favoring models that only detect larger lesions, a limitation often seen with standard DSC. By evaluating models on a lesion-by-lesion basis, we can better understand their ability to segment multi-focal and multi-centric disease.

\

In terms of the assessed and evaluated tumor sub-regions:
\begin{enumerate}
    \item ET describes the regions of active tumor as well as nodular areas of enhancement.
    \item NETC denotes necrosis and cysts within the tumor.
    \item SNFH includes edema, infiltrating tumor, and post-treatment changes.
    \item RC consists of both recent and chronic resection cavities and typically contains fluid, blood, air, and/or proteinaceous materials.
    \item Tumor core (ET plus NETC) describes what is typically resected during a surgical procedure.
    \item Whole tumor (ET plus SNFH plus NETC) defines the whole extent of the tumor, including the tumor core, infiltrating tumor, peritumoral edema and treatment-related changes.
\end{enumerate}

For ranking of multidimensional outcomes (or metrics), for each team, we will compute the summation of their ranks across the average of the metrics described above as a univariate overall summary measure. This measure will decide the overall ranking for each specific team. To visualize the results in an intuitive fashion, we propose to visualize the outcome via an augmented version of radar plot \cite{duan2023palm}.

\

Similar to BraTS 2017-2023 challenges, uncertainties in rankings will be assessed using permutational analysis \cite{bakas2018identifying}. Performance for the segmentation task will be assessed based on the relative performance of each team on each tumor sub-region and for each segmentation measure. These will be combined by averaging ranks for the measures, and statistical significance will be evaluated only for the segmentation performance measures and will be quantified by permutation the relative ranks for each segmentation measure and sub-region per subject of the testing data.
        \section{Discussion}

Assessment of diffuse glioma treatment response is challenging given the complex appearance on MRI after treatment. Automated segmentation of post-treatment diffuse glioma has significant potential to increase workflow efficiency with more rapid and accurate volumetric assessments for treatment planning \cite{rudie2022longitudinal, vollmuth2023artificial}. The 2024 BraTS post-treatment glioma challenge differs from previous BraTS glioma challenges as it consists of an entirely new dataset of exclusively post-treatment diffuse gliomas and includes a novel tissue class, the RC. The aim of the 2024 BraTS post-treatment glioma challenge will be to identify current state-of-the-art segmentation algorithms for post-treatment gliomas, and hence target a more clinically relevant question of monitoring tumor progression.

\

Prior studies in post-treatment glioma have shown slightly lower performance than for pre-treatment glioma regarding segmentations of large areas of abnormal FLAIR signal and areas of ET \cite{kickingereder2019automated, chang2019automatic, rudie2022longitudinal, lotan2022development, ghaffari2022automated, sorensen2023evaluation}. However, smaller areas of contrast enhancement still remain challenging, and many of these studies have not included the RC as a separate label. Providing labeled segmentations of the RC is important for correct therapy planning, especially for patients receiving adjuvant radiotherapy after surgical resection \cite{niyazi2023estro, ermics2020fully}.

\

Automated segmentation of post-treatment diffuse glioma tumor subregions has significant potential to increase workflow efficiency for radiologists, neuro-oncologists, radiation oncologists, and neurosurgeons\cite{kickingereder2019automated, vollmuth2023artificial, rudie2019emerging}. Given the significant effort required to annotate multilabel tissue classes in the post-treatment setting, the release of this dataset, which includes expert-voxelwise segmentations of tumor subregions, will support future studies seeking to validate automated segmentation tools for post-treatment diffuse gliomas.

\

Although efforts were made during challenge design and dataset preparation there are limitations to this study that should be mentioned. The annotator-approval model for manual segmentation corrections, while consistent with prior BraTS challenges \cite{baid2021rsna}, does not address inter-observer variability. To mitigate this variability, annotators received written instructions and an expert reviewer evaluated all segmentations prior to their inclusion in the dataset. To quantify this variability in this dataset, the test set was annotated and approved by two sets of annotators/approvers. 

\

There are several ways to improve and expand future BraTS glioma challenges.  The annotation protocol combines abnormal signals related to residual/recurrent tumor and post-treatment changes due to overlapping imaging features. Confidently distinguishing these entities is a challenging clinical task, which becomes even more difficult in the context of this dataset without relevant treatment history, prior imaging, or advanced imaging (diffusion and perfusion). Thus, future challenges would need to incorporate longitudinal and multimodal data in order to tackle the crucial issue of distinguishing between residual/recurrent enhancing and infiltrative tumor from edema, gliosis, granulation tissue and pseudoprogression. Such a model could significantly aid in developing optimal patient management plans.

\

Another challenge lies in identifying specific imaging features that can accurately anticipate responses to different treatments. Given its non-invasive nature, MRI is widely used in neuro-oncology practice for response evaluation, typically by measuring tumor size before and after treatment. However, because response patterns can be complex and heterogeneous, this approach does not always lead to an accurate assessment of the underlying biological response. Creating a model tailored for this task would require adding more clinical information to the dataset. By developing such a model, clinicians could potentially gain valuable insights into the disease's progression and tailor treatment strategies accordingly.

        \section{Conclusion}

The objective of the 2024 BraTS post-treatment glioma challenge is to establish a benchmark and define a community standard for automated segmentation on post-treatment MRI, utilizing the largest, publicly available, expert-annotated post-treatment glioma MRI dataset. The developed state-of-the-art models will provide a crucial tool for objectively assessing residual tumor volume for follow-up examinations and treatment planning that has the potential to improve patient management and outcomes. Additionally, they will lay the foundation for future studies aimed at identifying tumor subtypes, assessing aggressiveness, and predicting recurrence risk based solely on MRI findings.

    \section*{Acknowledgments}
        The success of any challenge in the medical domain depends on the development of a large and well-curated dataset. We express our gratitude to all data contributors, annotators, and approvers for their dedication and effort. 
    
    
    \bibliographystyle{ieeetr}
    \bibliography{bibliography.bib}
    \newpage
    \appendix
\end{document}